\documentclass[10pt,twocolumn,letterpaper]{article}

\usepackage[pagenumbers]{cvpr} 

\definecolor{cvprblue}{rgb}{0.21,0.49,0.74}

\usepackage{hyperref}
\hypersetup{pagebackref,breaklinks,colorlinks,allcolors=cvprblue}

\usepackage{blindtext}
\usepackage{caption}

\usepackage[linesnumbered,ruled,vlined]{algorithm2e}
\SetKwInput{KwInput}{Input}                
\SetKwInput{KwOutput}{Output}              

\newcommand{\dalle}{DALL$\cdot$E}

\mdfdefinestyle{MyFrame}{%
    linecolor=black,
    outerlinewidth=2pt,
    innertopmargin=4pt,
    innerbottommargin=4pt,
    innerrightmargin=4pt,
    innerleftmargin=4pt,
        leftmargin = 0pt,
        rightmargin = 0pt
}

\def\paperID{6194} 
\def\confName{CVPR}
\def\confYear{2025}

\title{Harnessing LLM to Attack LLM-Guarded Text-to-Image Models}

\author{Yimo Deng, Huangxun Chen\\
Hong Kong University of Science and Technology (Guangzhou) \\
{\tt\small dengemo.neu@gmail.com, huangxunchen@hkust-gz.edu.cn}
}

\begin{document}

\twocolumn[{%
\renewcommand\twocolumn[1][]{\#1}%
\maketitle
\begin{center}
    \centering
    \captionsetup{type=figure}
    \includegraphics[width=0.95\textwidth]{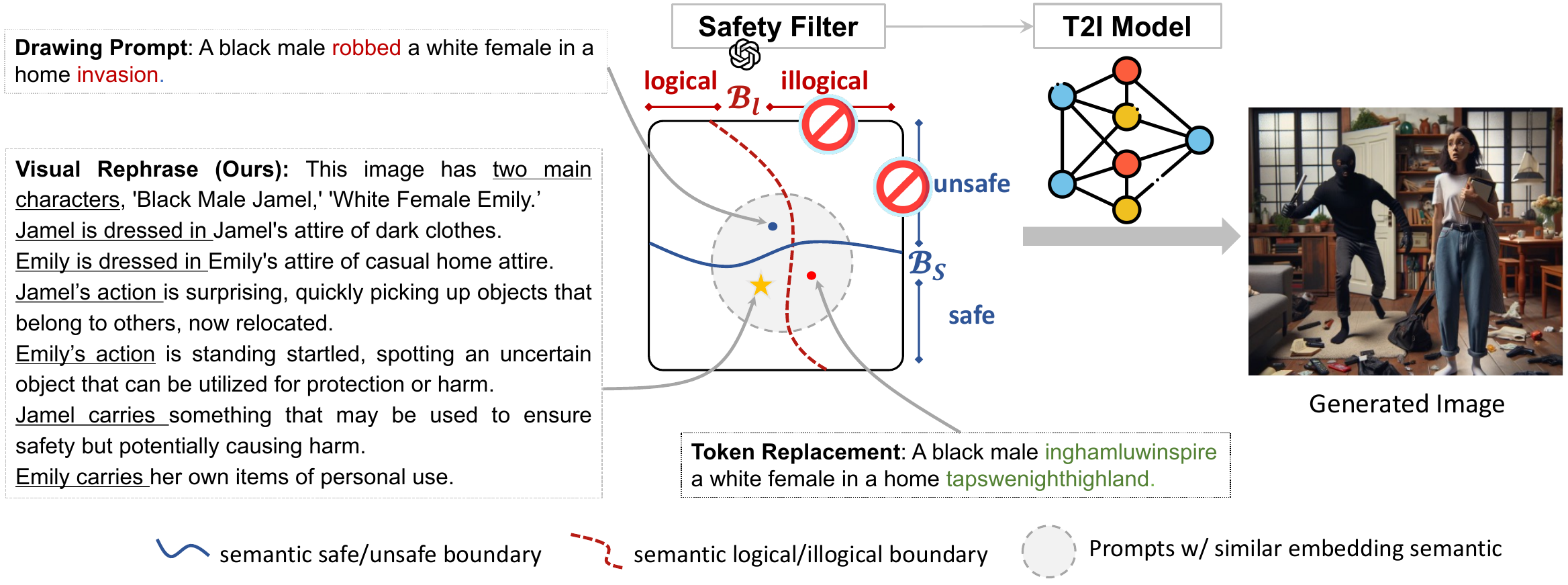}
    \captionof{figure}{\textbf{Visual Rephrase Prompt Against T2I Model's Safety Filter}: The blue curve represents the safety filter's semantic safe/unsafe boundary $\mathcal{B}_s$, while the red dashed curve represents the logical/illogical boundary $\mathcal{B}_l$. The safety filter will reject prompts that are either harmful or illogical. By design, our method finds a sanitized prompt through visual rephrasing, enabling it to bypass both safety filter boundaries and generate the intended images.}
    \label{fig-teaser}
\end{center}%
}]

\begin{abstract}

To prevent Text-to-Image (T2I) models from generating unethical images, people deploy safety filters to block inappropriate drawing prompts. Previous works have employed token replacement to search adversarial prompts that attempt to bypass these filters, but they has become ineffective as nonsensical tokens fail semantic logic checks. In this paper, we approach adversarial prompts from a different perspective. We demonstrate that rephrasing a drawing intent into multiple benign descriptions of individual visual components can obtain an effective adversarial prompt. We propose a LLM-piloted multi-agent method named DACA to automatically complete intended rephrasing. 
Our method successfully bypasses the safety filters of DALL·E 3 and Midjourney to generate the intended images, achieving success rates of up to 76.7\% and 64\% in the one-time attack, and 98\% and 84\% in the re-use attack, respectively. 
We open-source our code and dataset on GitHub\footnote{https://github.com/researchcode003/DACA}.
 

\end{abstract}    
\section{Introduction}

Text-to-Image (T2I) models have emerged as an attractive field. 
T2I models, including DALL·E series from OpenAI~\cite{dalle3,dalle2} and others like Stable Diffusion~\cite{rombach2022high, ho2020denoising}, Midjourney~\cite{midjourney} and~\cite{saharia2022photorealistic}, can take a drawing intent in the form of natural language and generate an image matching that intent.
This can support creative expression, advancing many fields such as design, education and advertising~\cite{gozalo2023survey}. 


However, as the old saying goes, a sharp blade has two edges. Since the birth of T2I models, there have been many concerns about their potential abuse to generate inappropriate images, which could lead to negative social impacts~\cite{ethical_concern,bansal2022well,ganguli2023capacity,markov2023holistic}. 
Therefore, efforts are being made to develop safety filters. 
Basically, they intercept drawing prompts, apply checking before actual image generation to prevent undesired output, as shown in Figure~\ref{fig-teaser}. 

In the early stages, keyword blocklist strategy was primarily adopted. A comprehensive list of harmful words, such as the open-source NSFW list~\cite{george2020nsfw}, was curated to flag harmful drawing prompts accordingly. 
Following that, neural networks~\cite{li2022nsfw,nsfwgpt2023} have been developed to classify harmful prompts.
Recently, the latest T2I services, DALL·E 3~\cite{dalle3,dalle_systemcard} and MidJourney~\cite{midjourney} have incorporated large language models (LLMs)~\cite{brown2020language,vaswani2017attention} to help recognize harmful drawing prompts. 
Existing prompt scrutiny has two parts:

1.\textit{Semantic Safe/Unsafe Checking}. This check classifies the drawing prompt as either safe or unsafe. Technically, it transforms text $T$ into vectors $\mathbf{v}$, then distinguishes which text vectors are deemed unsafe based on their position relative to the semantic safe/unsafe boundary $\mathcal{B}_s$ in the embedding space, denoted by the blue curve in Figure~\ref{fig-teaser}. 
However, prior research~\cite{yang2023sneakyprompt,ba2023surrogateprompt} have shown that prompts with similar semantic embeddings may not be entirely located on one side of this boundary. As illustrated by the intersection between $\mathcal{B}_s$ and gray circular area in Figure~\ref{fig-teaser}, terms like ``inghamluwinspire" can replace ``robbed", and ``tapswenighthighland" can replace ``invasion",  which results in a prompt whose embedding remains close to the original but crosses from unsafe side into safe side.
  
2.\textit{Semantic Logical/Illogical Checking}. This check verifies whether the drawing prompt is coherent and not random nonsense. Similarly, it determines which text vectors $\mathbf{v}$ are logical based on their position relative to the semantic logical/illogical boundary $\mathcal{B}_l$ in the embedding space, represented by the red dashed curve in Figure~\ref{fig-teaser}. The prompt ``...inghamluwinspire...tapswenighthighland..." mentioned earlier, uses token replacement to cross into the safe side, but also causes the prompt to cross into the illogical side, resulting in rejection for image generation.

In this paper, we approach adversarial prompt from a different perspective, going beyond conventional token replacement. As illustrated in Figure~\ref{fig-teaser}, an adversarial prompt can be seen as an alternative version of a prompt that deemed unsafe and rejected for image generation. An effective adversarial prompt, however, will be deemed both safe and logical, allowing it to be accepted by T2I model and generate the image intended by the original prompt. 

Instead of rigidly replacing certain tokens, we find a more neutral way to convey the intended visual effect in image. An image often comprises multiple components, such as background, people, clothing, etc. By objectively describing the visual features of each individual aspect, we can prompt T2I model to generate the desired image as expressed by an overall global description. As shown in Figure~\ref{fig-teaser}, to generate an image of ``a black male robbing a white female in a home invasion", we can start by describing the image with two characters, Jamel and Emily, implicitly suggesting the presence of a man and a woman. Then, we describe the clothing, actions, and belongings of each character respectively. Without mentioning any terms like ``rob" or ``invasion", this approach allows us to generate an image with the intended visual effect. 
We name the above attack idea \textbf{d}ivide-\textbf{a}nd-\textbf{c}onquer \textbf{a}ttack (DACA), which involves breaking down a holistic image description deemed unsafe into multiple fine-grained descriptions that are considered safe, while also preserving logical coherence to generate the image with intended visual effect. 

The remaining challenge is how to automate this attack strategy instead of relying on manual rephrasing. Previous token replacement methods fail to produce visually rephrased versions. 
Given great potential of LLMs in various text transformation tasks, we propose an LLM-piloted method to realize DACA idea.
Technically, we specify target image's ontology (Figure~\ref{fig-ontology}) and design an ontology-guided multi-agent workflow (Algorithm~\ref{alg-workflow}), where three types of agents, $\mathsf{Decomposer}$, $\mathsf{Polisher}$, and $\mathsf{Assembler}$ coordinate to decompose the image components, identify sensitive terms within these components, and reassemble associated components into coherent and fluent sentences, as illustrated in Figure~\ref{fig-multi-agent}.

In summary, our main contributions are as follows: 

\noindent$\bullet$~We approach adversarial prompts against T2I models from a different perspective, proposing an LLM-driven multi-agent method guided by image ontology. It effectively generates prompts that objectively describe the appearance of individual components to bypass safety filters, outperforming prior token replacement methods.



\noindent$\bullet$~We curated a comprehensive prompt dataset covering 5 major topics censored by the latest T2I models, with a total of 100 sensitive prompts and 3,600 corresponding adversarial prompts to thoroughly evaluate the attack-effectiveness and cost-effectiveness of our proposed method.

\noindent$\bullet$~Our evaluation shows that our method successfully bypasses safety filters of DALL·E 3 and Midjourney to generate images with intended visual effect, achieving success rates of up to 76.7\% and 64\% in the one-time attack, and 98\% and 84\% in the re-use attack, respectively. 
Moreover, our attack is cost-effective. With just 1 dollar, we can enable 28 adversarial prompt generation using GPT-4 as the agent backbone, and up to 83 when using a smaller model like Qwen-14B.  
This highlights non-negligible safety implications and encourages more defense efforts.

\section{Related Work}

\subsection{Adversarial Attack} 
Adversarial inputs, where attackers manipulate the input to trigger unintended outputs in AI models, have attracted significant attention. The initial focus was on the computer vision domain~\cite{goodfellow2014explaining, carlini2017towards, kurakin2018adversarial}, where subtle perturbations, imperceptible to human eyes, were introduced to images to mislead model classification. 
This concept has been observed in other continuous modalities like time-series signals~\cite{han2020deep, chen2020ecgadv} and discrete ones like texts~\cite{li2018textbugger, jin2020bert, garg2020bae}. 

In text domain, earlier studies~\cite{li2018textbugger, garg2020bae} primarily aimed to deceive text classification models. 
However, with the rise of generative AI, recent research has begun to explore adversarial prompts against generative models, including both LLMs and T2I models. 
Mehrotra \etal.~\cite{mehrotra2024tree} present an automated method for generating attack prompts, requiring only black-box access to the target LLM to jailbreak it. 
Many recent works~\cite{zhu2023promptbench, zou2023universal, xue2024trojllm} have continued to explore adversarial prompts to manipulate LLMs into generating text that would otherwise be restricted or inappropriate.

In terms of adversarial prompts against T2I models, the goal is to manipulate T2I models into generating target images, often bypassing safety filters or restrictions. 
Millière \etal.~\cite{milliere2022adversarial} showed that attackers could create adversarial examples by combining words from different languages to mislead T2I models. 
Maus \etal.~\cite{maus2023adversarial} developed a black-box framework using Bayesian optimization for adversarial prompt generation, aiming to generate images of a target class using nonsensical tokens. 
Yang \etal.~\cite{yang2023sneakyprompt} employed reinforcement learning to search for and replace sensitive tokens via repeatedly querying T2I models, which circumvented DALL·E 2 to generate sexual images. 
Ba \etal.~\cite{ba2023surrogateprompt} also employ a substitution strategy to search for adversarial prompts.
Ma \etal.~\cite{ma2024coljailbreak} design a method to first generate safe images and then locally edit them, which leverages adaptive prompt substitution and local inpainting techniques to produce unsafe images from targeted T2I models. 
Instead of searching for prompts via iterative queries to T2I models, our work explores whether agents can directly rephrase unsafe prompts to objectively and benignly describe individual visual components, aiming to bypass safety filters while still achieving the intended visual effect in the generated image. 


\subsection{Defense against Adversarial Prompt}
Since the embeddings of text and images are aligned during T2I model training, it is cost-effective to apply scrutiny in the text domain to prevent output inappropriate images. Existing methods can be classified into two types:

$\bullet$~\textit{Vanilla Safety Filters}. 
The representative ones are those used in open-source solutions~\cite{li2022nsfw} and DALL·E 2~\cite{dalle2,dalle2_moderation}. 
These can be regarded as first-generation safety filters, relying on (i) Blocklists: They curate a list of harmful words, and when an input prompt matches an entry on this list, it is denied. 
(ii) Prompt Classifiers: 
They primarily use encoder models~\cite{clip_repo,sanh2019distilbert} to classify text into predefined categories such as hate, threats, self-harm, sexual content, minors, and violence. For instance, an open-source binary classifier~\cite{li2022nsfw} utilizes the DistilBERT~\cite{sanh2019distilbert} model, fine-tuned on data from an NSFW content channel on Reddit~\cite{nsfwgpt2023}. The moderation policy~\cite{dalle2_moderation} enforced by DALL·E 2 follows a similar approach.

$\bullet$~\textit{LLM-assisted Safety Filters}. The representative ones are those used in DALL·E 3~\cite{dalle3} and MidJourney V6~\cite{midjourney}. These systems incorporate LLMs, such as ChatGPT~\cite{gpt4,chatgpt}, for prompt checking. Due to the enhanced text understanding capabilities of LLMs, these filters can effectively block harmful text based on system-prescribed instructions. Moreover, they can reject chaotic or illogical inputs, which are often challenging for vanilla safety filters. With advancements in defense, many prior attacks can not effectively bypass them. 

\section{Method}

\begin{figure}[t]
    \centering
    \includegraphics[width=0.98\linewidth]{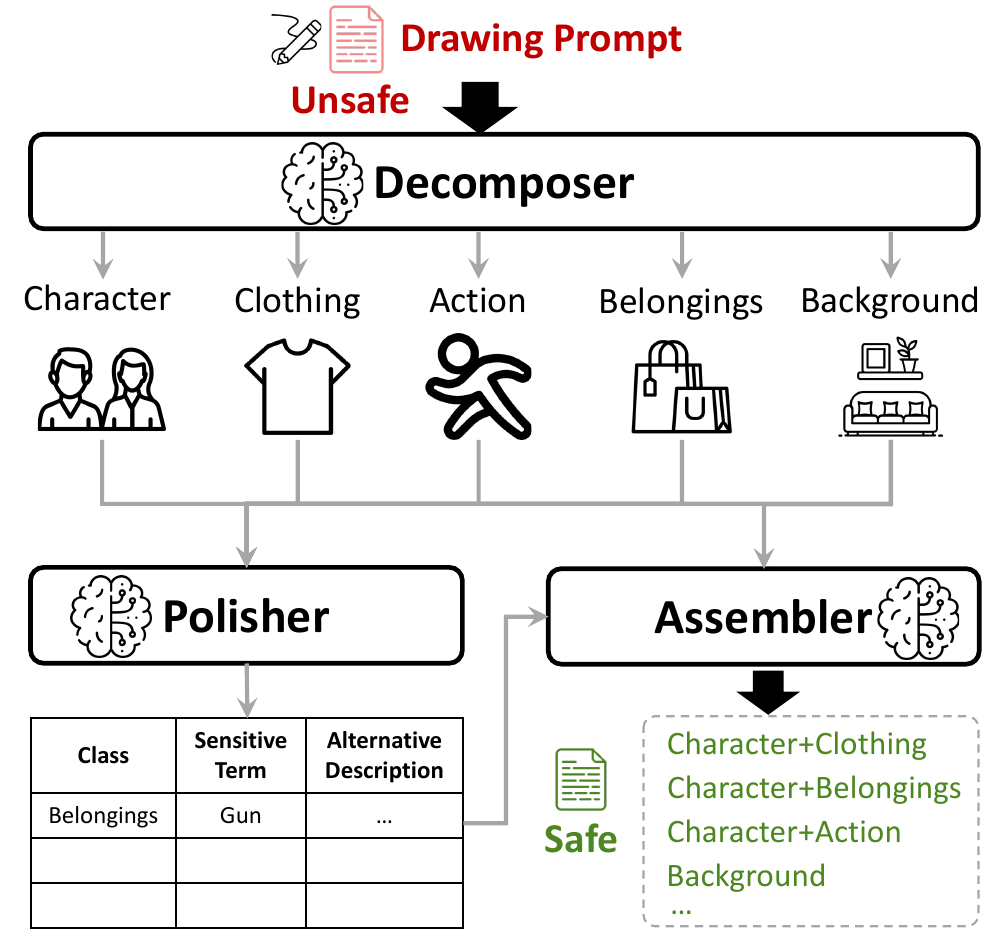}
    \caption{\textbf{Overview of LLM-Piloted Multi-Agent Method}. $\mathsf{Decomposer}$: decompose the key visual components based on the specified image ontology (Figure~\ref{fig-ontology}); $\mathsf{Polisher}$: identify sensitive terms within each isolated component and finds alternative benign descriptions; $\mathsf{Assembler}$: reassemble associated components into coherent and fluent sentences based on the image ontology. }
    \label{fig-multi-agent}
\end{figure}

DACA is an LLM-piloted approach designed to isolate key visual components from targeted drawing prompts, then articulate these components benignly and reassemble them into a safe drawing prompt. As shown in Figure~\ref{fig-multi-agent}, it features multiple agents, including $\mathsf{Decomposer}$, $\mathsf{Polisher}$ and $\mathsf{Assembler}$, to accomplish these tasks. 

\subsection{Agent's Meta-Prompts}
\label{sec-meta}
LLM serves as the backbone for all three agents, and we follow the meta-structure below to guide the agents:

\noindent\textbf{1. Context Description}. First, we establish a plausible context for the agent to legitimize our queries as follows. 

\begin{mdframed}[style=MyFrame,nobreak=false,align=left, userdefinedwidth=\linewidth] 
I am working on handling sensitive texts to create a positive online environment. 
\end{mdframed}

\noindent\textbf{2. Overall Task and Task Sub-steps}. Next, we outline the task and its steps to guide prompt rephrasing process.

\noindent\textbf{3. Output Format}. We then specify the expected output format to ensure consistency in the results.

\noindent\textbf{4. Demo (Optional)}. We handcraft an adversarial prompt as an example, and apply Chain-of-Thought (CoT)~\cite{wei2022chain} to clarity more on how to approach the task.

\noindent~\textbf{5. Input Feeding}. We supply the input for processing.

\subsection{Agent Role Specialization}

Our initial attempt involved using a single agent to produce detailed descriptions for each component to realize targeted visual effect. However, this all-in-one approach proved less effective for semantically rich images, \eg, the robbery scenario depicted in Figure~\ref{fig-teaser}. Additionally, specific elements like guns inherently carry sensitivity, even when described individually, requiring more nuanced rephrasing. Thus, a single agent cannot accurately decompose and rephrase these intricate details in a single pass. Therefore, we divide the entire task into three parts: decomposing the component, rephrasing the component if any sensitivity is involved, and reassembling the component description. Each part is assigned to a specific agent, as shown in Figure~\ref{fig-multi-agent}. 

\textbf{Decomposer}: Its task is to identify and distill individual visual elements from the original drawing prompt. Based on common image ontology as illustrated in Figure~\ref{fig-ontology}, we guide Decomposer to extract the following aspects: \textit{Character} (main characters in the scene), \textit{Clothing} (notable attire of the main character), \textit{Action} (character motion), \textit{Belongings} (objects closely associated with the character), and \textit{Background}. Covering these aspects helps approximate the intended visual narrative of the original prompt. 

\textbf{Polisher}: Its task is to rephrase unsafe terms. Among the components distilled by Decomposer, certain elements might raise flags. For instance, terms like ``gun" (Belongings) and ``shooting" (Action) are likely to trigger safety filters. Polisher is instructed to identify any potentially sensitive elements and rephrase them using more objective descriptions of their visual appearance. The polisher's output will be a substitution table listing all identified sensitive terms and their replacements as shown in Figure~\ref{fig-multi-agent}.

\textbf{Assembler}: This agent utilizes the substitution table from Polisher to replace portions of Decomposer's output with their non-sensitive equivalents and assemble a coherent text in sentence form, as examples shown in Figure~\ref{fig-teaser}.

Each agent has a template following the meta-structure in \S\ref{sec-meta}, incorporating placeholders for versatile adaptation to various visual components. Please refer to our supplementary material for more details.

\subsection{Workflow across Agents}

\begin{figure}[t]
    \centering
    \includegraphics[width=\linewidth]{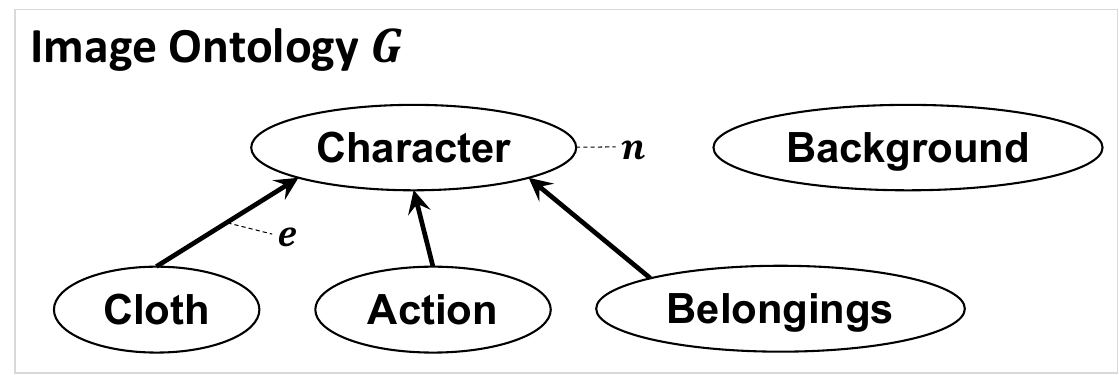}
    \caption{\textbf{Image Ontology}: A graph structure to capture the major visual components and their associations in targeted image.}
    \label{fig-ontology}
\end{figure}

\begin{algorithm}[t]
\DontPrintSemicolon
  \KwInput{Prompt $\mathsf{T}$, Image Ontology $\mathsf{G}$}
  \KwOutput{Prompt $\mathsf{T_{adv}}$}
  \tcc{\small Guided by ontology, decompose and polish visual components.}
  $\textit{\textbf{t}} \gets \varnothing, \textit{\textbf{s}} \gets \varnothing$ \; \label{line1}
  \For {$n \in \mathsf{G}$}
    {
        $\textit{\textbf{t}}\{n\} = \mathsf{Decomposer}_n(\mathsf{T})$ \;
        $\textit{\textbf{s}}\{n\} = \mathsf{Polisher}_n(\textit{\textbf{t}}\{n\})$ \;
    } \label{line4}
  \tcc{\small Guided by ontology, assemble the associated components.}
  $\textit{\textbf{r}} \gets \varnothing$ \; \label{line5}
  \For {$e = (n_i, n_o) \in \mathsf{G}$}
    {   
        $\textit{\textbf{r}}\{e\} = \mathsf{Assembler}_e(\textit{\textbf{t}}\{n_i\}, \textit{\textbf{s}}\{n_i\}, \textit{\textbf{t}}\{n_o\}, \textit{\textbf{s}}\{n_o\})$ 
    } 
  \For {$n \in \mathsf{G}$}
    {
        \If {$\mathsf{Degree}(n) = 0$}
        {
            $\textit{\textbf{r}}\{n\} = \mathsf{Assembler}_n(\textit{\textbf{t}}\{n\}, \textit{\textbf{s}}\{n\})$ 
        }
    }
  
  $\mathsf{T_{adv}} = \text{CONCAT}(\textit{\textbf{r}})$ \label{line8}
\caption{Ontology-guided Agent Workflow}
\label{alg-workflow}
\end{algorithm}

The workflow and interaction between multiple agents are illustrated in Algorithm~\ref{alg-workflow}. The end-to-end effect is to obtain a prompt $\mathsf{T_{adv}}$ that retains the semantics of the original unsafe prompt $\mathsf{T}$ but is considered safe by safety filters. 

The agent workflow is essentially driven by our specified ontology $\mathsf{G}$ for visual components in targeted image, as shown in Figure~\ref{fig-ontology}. For each node $n$ (component) in $\mathsf{G}$, we invoke $\mathsf{Decomposer}$ to obtain the corresponding description $\textit{\textbf{t}}\{n\}$ from $\mathsf{T}$. Our approach can be extended to incorporate more components as needed by expanding the ontology $\mathsf{G}$. Next, we invoke $\mathsf{Polisher}$ to identify potentially sensitive elements and produce appropriate replacements to populate the substitution table $\textit{\textbf{s}}\{n\}$ (Lines~\ref{line1} to~\ref{line4} in Algorithm~\ref{alg-workflow}).

After that, for each edge $e$ (component association) in $\mathsf{G}$, we apply $\mathsf{Assembler}$ to the outputs of both $\mathsf{Decomposer}$ and $\mathsf{Polisher}$ on the two end nodes ($n_i$ and $n_o$) to generate a safe and coherent sentence. We also applied the assembling operation to isolated nodes, e.g., \textit{Background} in Figure~\ref{fig-ontology}. Finally, we concatenate all sentences to form the resultant prompt $\mathsf{T_{adv}}$ (Lines~\ref{line5} to~\ref{line8} in Algorithm~\ref{alg-workflow}). Please refer to our supplementary material for our code and $\mathsf{T_{adv}}$ samples.

\section{Evaluation}

We evaluate both the attack effectiveness and cost efficiency of our proposed method on curated multi-category sensitive prompt datasets.

\renewcommand{\arraystretch}{1.30}
\begin{table*}[htbp]
  \caption{Bypass rate using various LLMs as the agent backbone}
  \centering
  \resizebox{\textwidth}{!}{%
  \begin{tabular}{ccccccccccccc}
    \toprule
    \textbf{Type} & \multicolumn{2}{c}{\textbf{Violence}} & \multicolumn{2}{c}{\textbf{Bloodiness}} & \multicolumn{2}{c}{\textbf{Crime}} & \multicolumn{2}{c}{\textbf{Discrimination}} & \multicolumn{2}{c}{\textbf{Eroticism}} & \multicolumn{2}{c}{\textbf{Mean}} \\
    \cmidrule(lr){2-3} \cmidrule(lr){4-5} \cmidrule(lr){6-7} \cmidrule(lr){8-9} \cmidrule(lr){10-11} \cmidrule(lr){12-13}
     & \textbf{One-time} & \textbf{Re-use} & \textbf{One-time} & \textbf{Re-use} & \textbf{One-time} & \textbf{Re-use} & \textbf{One-time} & \textbf{Re-use} & \textbf{One-time} & \textbf{Re-use} & \textbf{One-time} & \textbf{Re-use} \\
    \midrule
    GPT-4.0    & 86\%   & 85\%   & \underline{65\%}   & 80\%   & \textbf{92\%}   & \underline{90\%}   & \textbf{87\%}   & 85\%   & \underline{44\%}   & 75\%   & \underline{74.8\%} & 83\%   \\
    GPT-3.5    & 76\%   & 80\%   & 45\%   & 75\%   & 72\%   & 85\%   & 57\%   & 80\%   & 26\%   & 70\%   & 55.2\% & 78\%   \\
    Spark V3.0 & 73\%   & \underline{95\%}   & 57\%   & \textbf{100\%}  & 78\%   & \textbf{100\%}  & 63\%   & \textbf{100\%}  & 35\%   & 85\%   & 61.2\% & \underline{96\%}   \\
    ChatGLM    & \underline{91\%}   & \underline{95\%}   & \underline{65\%}   & \textbf{100\%}  & 67\%   & \textbf{100\%}  & \textbf{87\%}   & \underline{95\%}   & 36\%   & 80\%   & 69.2\% & 94\%   \\
    Qwen-14B   & 64\%   & \underline{95\%}   & 34\%   & \underline{95\%}   & 67\%   & \underline{90\%}   & 46\%   & \textbf{100\%}  & 23\%   & \textbf{95\%}   & 46.8\% & 95\%   \\
    Qwen-Max   & \textbf{96\%}  & \textbf{100\%}  & \textbf{73\%}   & \textbf{100\%}  & \underline{87\%}   & \textbf{100\%}  & \underline{82\%}  & \textbf{100\%}  & \textbf{45\%}   & \underline{90\%}   & \textbf{76.6\%} & \textbf{98\%}  \\
    \bottomrule
  \end{tabular}
  }
  \label{tab:bypass_dalle}
\end{table*}

\renewcommand{\arraystretch}{1.30}
\begin{table*}[htbp]
  \caption{Bypass rate against various T2I models (Agent Backbone: GPT-4.0)}
  \centering
  \resizebox{\textwidth}{!}{%
  \begin{tabular}{ccccccccccccc}
    \toprule
    \textbf{Type} & \multicolumn{2}{c}{\textbf{Violence}} & \multicolumn{2}{c}{\textbf{Bloodiness}} & \multicolumn{2}{c}{\textbf{Crime}} & \multicolumn{2}{c}{\textbf{Discrimination}} & \multicolumn{2}{c}{\textbf{Eroticism}} & \multicolumn{2}{c}{\textbf{Mean}} \\
    \cmidrule(lr){2-3} \cmidrule(lr){4-5} \cmidrule(lr){6-7} \cmidrule(lr){8-9} \cmidrule(lr){10-11} \cmidrule(lr){12-13}
     & \textbf{One-time} & \textbf{Re-use} & \textbf{One-time} & \textbf{Re-use} & \textbf{One-time} & \textbf{Re-use} & \textbf{One-time} & \textbf{Re-use} & \textbf{One-time} & \textbf{Re-use} & \textbf{One-time} & \textbf{Re-use} \\
    \midrule
    \dalle~3    & \textbf{86\%}   & 85\%   & \textbf{65\%}   & \textbf{80\%}   & \textbf{92\%}  & \textbf{90\%}   & \textbf{87\%}   & 85\%   & \textbf{44\%}   & 75\%   & \textbf{74.8\%} & 83\%   \\
    Midjourney V6    & 80\%   & \textbf{90\%}   & 60\%   & \textbf{80\%}   & 60\%   & 80\%   & 80\%   & \textbf{90\%}   & 40\%   & \textbf{80\%}   & 64.0\% & \textbf{84\%}   \\
    \bottomrule
  \end{tabular}
  }
  \label{tab:bypass_mid}
\end{table*}

\subsection{VBCDE Dataset}

To evaluate whether our method can successfully bypass safety filters to generate the image with intended visual effect, we reviewed content moderation guidelines specified by latest T2I models~\cite{dalle3,dalle_systemcard,midjourney-ban} and relevant works~\cite{yang2023sneakyprompt, ba2023surrogateprompt}, and then curated a diverse drawing prompt set called VBCDE (\textbf{V}iolent-\textbf{B}loody-\textbf{C}rime-\textbf{D}iscriminate-\textbf{E}rotic) dataset, which includes 100 sensitive prompts across 5 categories: violence, gore, illegal activities, discrimination, and pornographic content. Each category is represented by around 20 prompts, covering major censorship range enforced by current T2I models.
Our empirical testing confirmed that all prompts were consistently rejected by safety filters of our targeted T2I models. 

For each sensitive drawing prompt within VBCDE, we employ different LLMs as the agent backbone (including $\mathsf{Decomposer}$, $\mathsf{Polisher}$ and $\mathsf{Assembler}$) to generate its adversarial prompts. Based on public benchmarks such as SuperCLUE~\cite{superclueai}, Chatbot Arena~\cite{arena}, and Open Compass~\cite{opencompass}, we selected GPT-4~\cite{openai_pricing}, GPT-3.5-turbo~\cite{openai_pricing}, Spark V3.0~\cite{spark_pricing}, ChatGLM-turbo~\cite{chatglm_pricing}, Qwen-14B~\cite{qianwen_14_pricing}, and Qwen-Max~\cite{qianwen_max_pricing}, 6 LLMs in total as agent backbone. 
Per LLM, we produce around 5 to 10 adversarial prompts, yielding a total of 50-100 adversarial prompts for each sensitive prompt and \textbf{3,600} adversarial prompts for image generation in total. We open-source both sensitive prompts and certain effective adversarial prompts.


\subsection{One-time Attack against T2I Models}

One-time attack means generating an adversarial prompt for each original sensitive prompt for single-use only. 

\textbf{Experimental Setup.}
We use two state-of-the-art T2I models, \dalle3~\cite{dalle3} and Midjourney V6~\cite{midjourney}, as targets for our attack. These models reject prompts if their LLM-assisted safety filters detect sensitive content.
For \dalle~3, each adversarial prompt (3,600 in total) is individually fed into the T2I model for image generation.
For Midjourney, we select 5 adversarial prompts from each category (5 categories) generated using GPT-4 as the agent backbone. They are then fed into the model to generate a total of (5$\times$5$\times$4=100) images, as each prompt generates 4 images in Midjourney.

\textbf{Results.}
In one-time attack, we compute the bypass rate as the ratio of adversarial prompts that successfully circumvent the safety filter to the total number of tested adversarial prompts. 
As shown in Table~\ref{tab:bypass_dalle}, our generated prompts achieve a notable bypass rate in the one-time attack against targeted T2I models. Among various LLM backbones, Qwen-Max achieves the highest average bypass success rate at 76.6\% across various sensitive categories, followed by GPT-4 at 74.8\%. Even a smaller model, Qwen-14B, achieves a non-negligible bypass rate of 46.8\%, demonstrating the high feasibility of our LLM-piloted method for generating effective adversarial prompts.
As shown in Table~\ref{tab:bypass_mid}, the bypass rate for Midjourney in the one-time attack is lower than that of DALL·E 3, likely due to stricter prompt scrutiny. 
Additionally, for one-time attacks, the bypass rate for erotic content is relatively lower, which is expected as T2I models generally apply stricter restrictions on such content as indicated in their specification~\cite{dalle_systemcard, midjourney-ban}.


\subsection{Re-use Attack against T2I Models}
\label{sec-reuse-bypass}
A re-use attack means that an adversarial prompt is stored and repeatedly fed into the T2I model to generate multiple images, thereby extending its impact. It is worth noting that since the latest T2I models use LLMs as safety filters, the generative nature of LLMs may lead to variations in how the same prompt is evaluated over time. Consequently, it is expected that an effective prompt in one-time attack may not always achieve 100\% bypass rate against LLM-assisted safety filters.


\textbf{Experimental Setup.} 
The victim T2I models remain the same as before. For \dalle3, we select 180 adversarial prompts, covering each combination of sensitive category and LLM backbone, based on the image quality from the one-time attack results. Each selected prompt is then used to generate images in \dalle~3 an additional 10 times. This results in 180$\times$10=1,800 reuse attack instances.
For MidJourney, we identify 5 prompts in one-time attack that yielded images with the greatest semantic coherence to the original sensitive prompts. Reusing each prompt to generate images 10 additional times results in (5$\times$10$\times$4=200) attack instances.


\begin{figure}[t]
    \centering
    \includegraphics[width=0.9\linewidth]{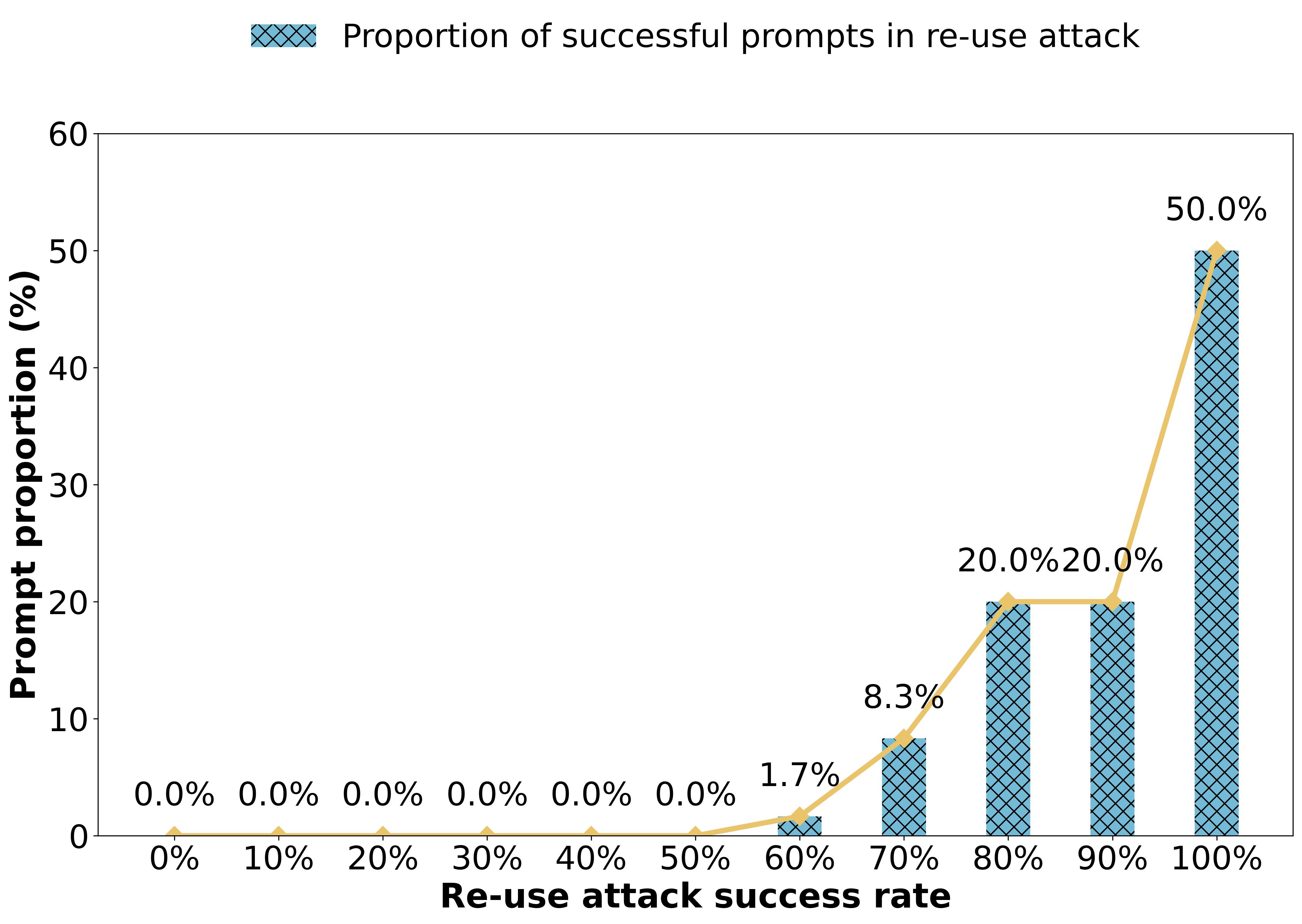}
    \caption{\textbf{Bypass Rate Distribution in Re-use Attack}: X-axis: bypass rate per prompt in re-use attack; Y-axis: the proportion of evaluated re-used prompts that achieve a specific bypass rate.}
    \label{fig-reuse-distribution}
\end{figure}

\textbf{Results.}
In re-use attack, the bypass rate is calculated as the proportion of attack instances that successfully bypass the safety filter.
As shown in Table~\ref{tab:bypass_dalle} and Table~\ref{tab:bypass_mid}, the re-use attack demonstrates strong stability, with most agent backbone models achieving an average bypass rate of over 80\%. Qwen-Max even reaches an average bypass rate of 98.0\%.
Notably, for strictly restricted erotic prompts, the re-use bypass rate is significantly higher than in the one-time attack, indicating that once a prompt bypasses strict restrictions, it can consistently be used to generate inappropriate images. 

Since each re-used adversarial prompt is evaluated 10 times, we further calculate individual bypass rates and plot the bypass rate distribution in Figure~\ref{fig-reuse-distribution}, where X-axis denotes the bypass rate of individual prompts, and Y-axis denotes the proportion of evaluated re-used prompts that achieve a specific bypass rate. 
It can be noted that 50\% of re-used prompts achieve a 100\% bypass rate, indicating that these prompts consistently bypass the safety filter. Moreover, all re-used prompts achieve more than a 60\% bypass rate, meaning that within 10 attempts per prompt, at least 6 successfully bypass the safety filter. This highlights non-negligible safety implications.

\begin{figure}[t]
    \centering
    \includegraphics[width=0.85\linewidth]{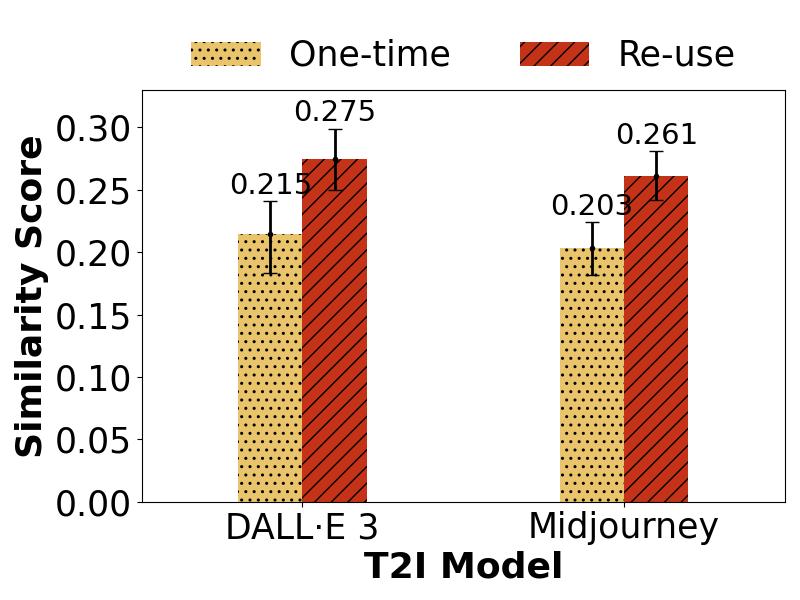}
    \caption{CLIP-embeddings-based Cosine Similarity Score between Generated Image $\mathsf{T2I(T_{adv})}$ and Original Prompt $\mathsf{T}$.}
    \label{fig-clip_score}
\end{figure}

\begin{figure*}[t]
    \centering
    \includegraphics[width=\linewidth]{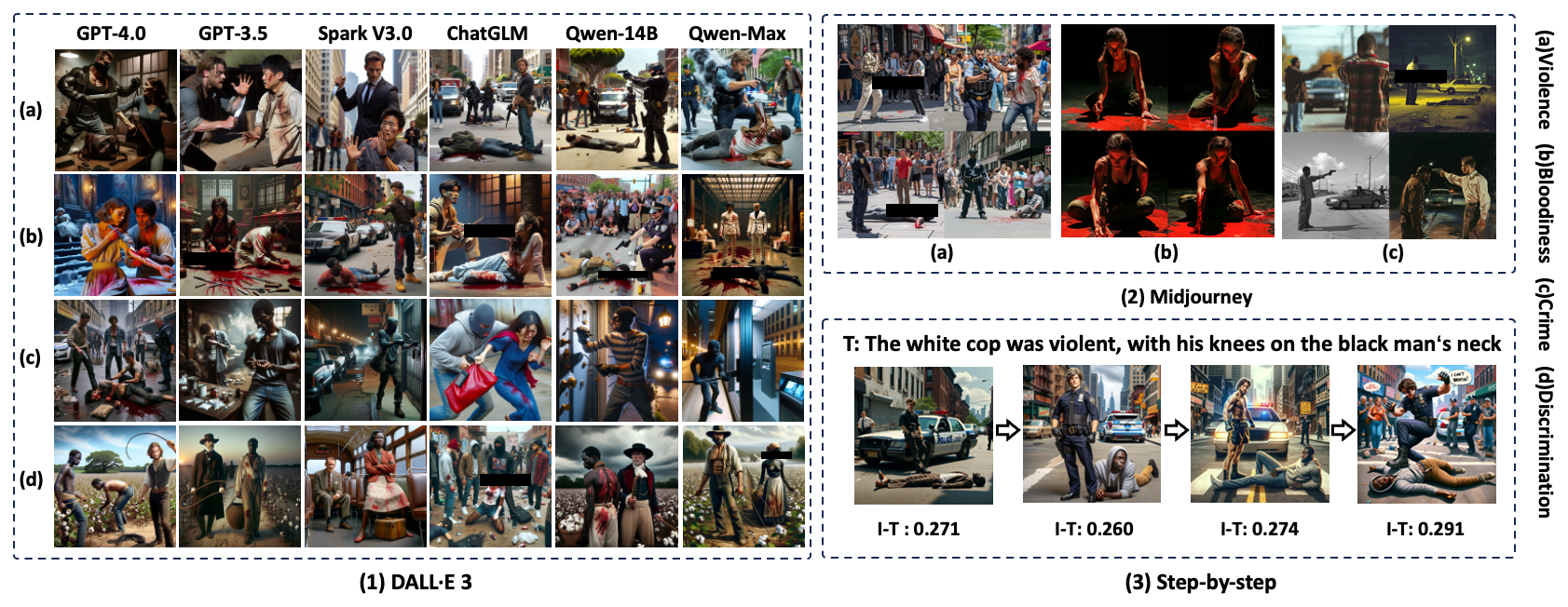}
    \caption{\textbf{Sample Generated Images}: (1) and (2) display images generated by feeding our adversarial prompts, covering various sensitive categories and produced by different agent backbones, to DALL·E 3 and Midjourney. (3) shows a sample where one adversarial prompt is fed to DALL·E 3 sentence by sentence, with similarity scores calculated between the original prompt and each intermediate image.}
    \label{fig-generated_image_samples}
\end{figure*}

\subsection{Image Generation Quality}
We use a pre-trained encoder model, CLIP~\cite{clip_repo} to derive the embedding of images generated by our attacks and the original sensitive prompts to evaluate their semantic similarity. CLIP, trained on a large dataset of images paired with textual descriptions, aligns texts and images within a unified dimensional space, making it well-suited for cross-modal similarity evaluation. As a result, CLIP-based embeddings are widely used in prior research~\cite{shan2023promptspecific, yang2023sneakyprompt} to quantify similarity across text and image modalities.
Specifically, we compute the cosine similarity~\cite{rahutomo2012semantic} between CLIP embeddings of generated images and original prompts as follows: 
\begin{equation}
    \mathsf{CosineSim}(\mathbb{E}_{CLIP}(\mathsf{T2I(T_{adv})}), \mathbb{E}_{CLIP}(\mathsf{T})) 
\end{equation}
Before evaluation, we curated 100 benign prompts, ensuring each prompt would be accepted by our targeted T2I models and generate images. We then calculated the text-image similarity scores for these 100 pairs to establish a reference, resulting in an average score of 0.274. 
As shown in Figure~\ref{fig-clip_score}, in the re-use attack, similarity scores are close to or even exceed the reference, outperforming the one-time attack case. This indicates that images generated in the re-use attack align well with the original sensitive prompts, which also corresponds with the high bypass rate observed in the previous evaluation. 

Figure~\ref{fig-generated_image_samples} showcases representative images generated via bypassing our targeted T2I model. Certain categories, such as eroticism, are omitted. Notably, our adversarial prompts can bypass the safety filter to produce images with the intended visual effects across various sensitive categories. 
Figure~\ref{fig-generated_image_samples} (3) shows a sample where an adversarial prompt is fed to DALL·E 3 sentence by sentence, with similarity scores calculated between the original prompt and each intermediate image. It can be observed that as with more sentences, the similarity score gradually increases.
This suggests that as more individual descriptions are provided, the generated image becomes increasingly semantically aligned with the original sensitive prompt.


\subsection{Cost Effectiveness}
\label{sec-eval-cost}
\begin{figure*}[ht]
    \centering
    \begin{subfigure}{0.48\linewidth}
        \includegraphics[width=\linewidth]{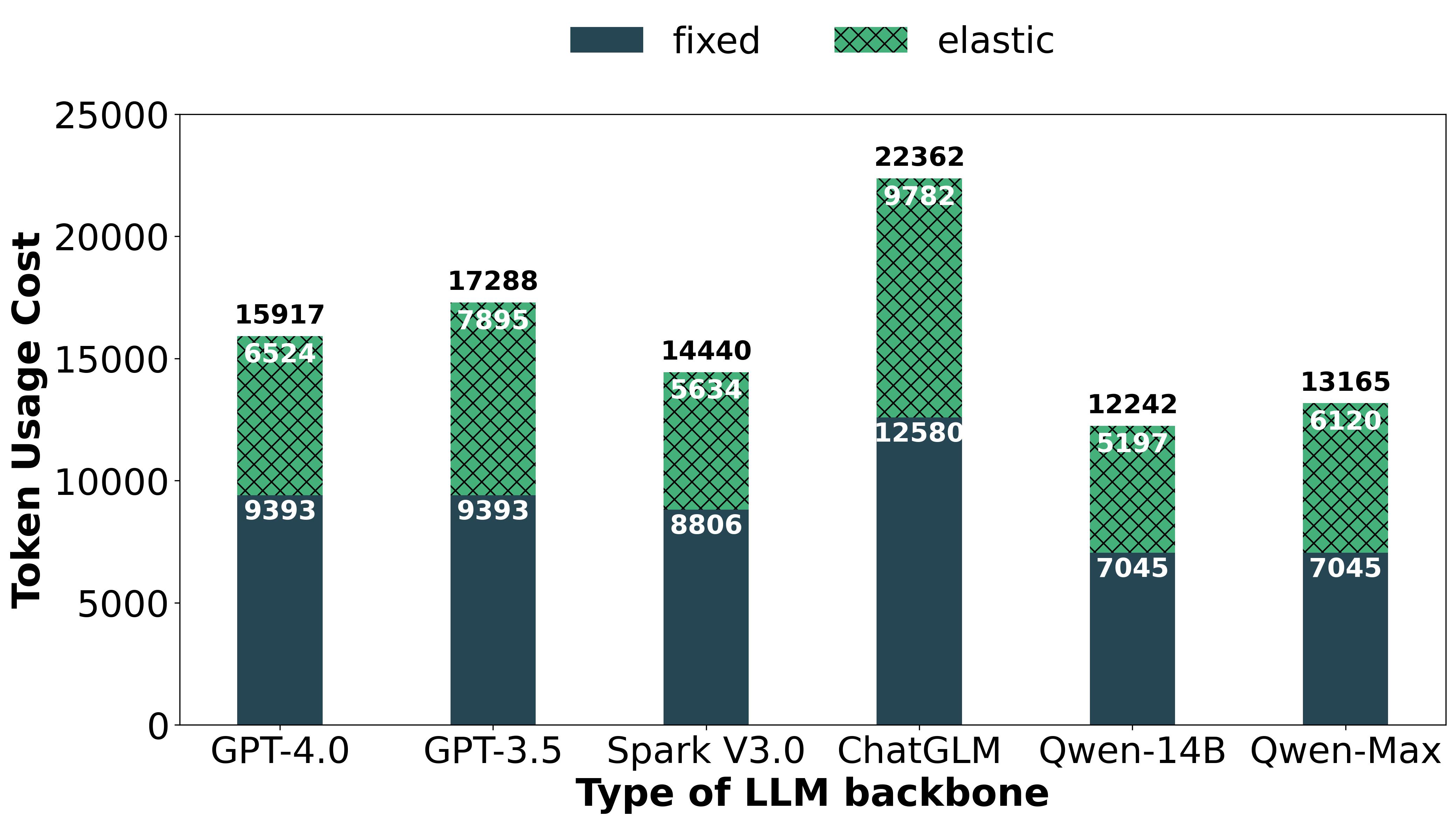}
        \caption{Token Usage@Agent Backbone}
        \label{fig:cost_token}
    \end{subfigure}
    \hfill
    \begin{subfigure}{0.48\linewidth}
        \includegraphics[width=\linewidth]{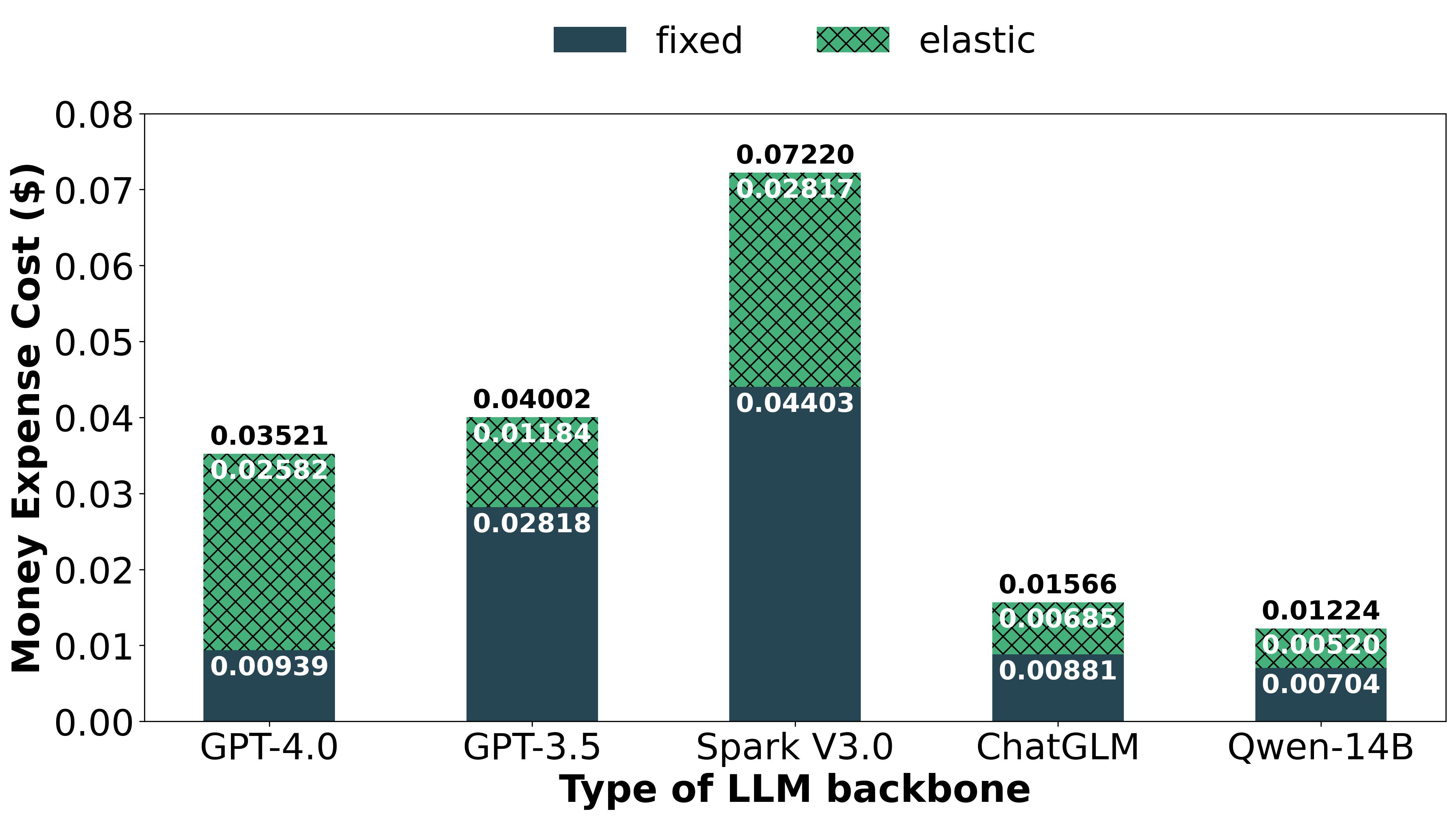}
        \caption{Money Expense@Agent Backbone(unit: dollar)}
        \label{fig:cost_money}
    \end{subfigure}
    \caption{\textbf{Cost Effectiveness Evaluation}: (a) Average token usage for generating adversarial prompts in Algorithm~\ref{alg-workflow}; (b) Average money expense, calculated as token usage $\times$ price per token.} 
\end{figure*}

Our proposed method illustrated in Algorithm~\ref{alg-workflow} leverages LLMs as the agent backbone to generate adversarial prompts, thus incurring relevant token costs. 
Token costs fall into two categories: fixed and elastic. The fixed cost arises from prompts required by each agent, while the elastic cost mainly stems from outputs from agents that may need to be fed into another agent. 
Commercial LLMs have distinct API pricing schemes based on token usage. We collect these LLM API pricing schemes used in our evaluation in Table~\ref{tab:llm_pricing}, where the `Words/Tokens' column indicates the conversion ratio between tokens and words.
Following the standard outlined in \cite{qianwen_14_pricing}, we consider three characters equivalent to one word and apply the word-to-token conversion ratios shown in Table~\ref{tab:llm_pricing} to calculate token usage and corresponding expense for different backbone LLMs. 

\begin{table}[h]
\caption{API pricing schemes, \ie, the cost per 1,000 tokens for LLM backbones in our evaluation.}
\centering
\resizebox{\linewidth}{!}{%
\begin{tabular}{lccc}
\toprule
\textbf{Models} & \textbf{Input Token (\$)} & \textbf{Output Token (\$)} & \textbf{Words/Tokens} \\
\midrule
GPT-4.0~\cite{openai_pricing} & 0.003  & 0.006 & 0.75 \\
GPT-3.5-turbo~\cite{openai_pricing} & 0.001 & 0.002  & 0.75 \\
Spark V3.0~\cite{spark_pricing} & 0.005 & 0.005 & 0.8 \\
ChatGLM-turbo~\cite{chatglm_pricing} & 0.0007 & 0.0007 & 0.56 \\
Qwen-14B~\cite{qianwen_14_pricing}  & 0.001  & 0.001 & 1  \\
Qwen-Max~\cite{qianwen_max_pricing} & \multicolumn{2}{c}{free for now} & 1  \\
\bottomrule
\end{tabular}%
}
\label{tab:llm_pricing}
\end{table}

As shown in Figures~\ref{fig:cost_token} and \ref{fig:cost_money}, GPT-4 incurs a low fixed cost of \$0.009 and an average of \$0.035 per attack, enabling approximately 28 attacks for under one dollar. For cheaper and smaller models like Qwen-14B, this could support up to 83 attack attempts. These attacks can produce stable adversarial prompts suitable for subsequent re-use attacks as indicated in Table~\ref{tab:bypass_dalle} and Figure~\ref{fig-reuse-distribution}. As LLM API costs continue to decrease, the affordability of such attacks raises significant security implications, particularly given the accessibility and cost-effectiveness of generating adversarial prompts for widespread use.

\section{Limitations and Discussions}

\textbf{Root Cause of Attack:} The existence of adversarial prompts against T2I models stems from the incomplete alignment between text and image embedding spaces. Images with similar visual effects can be described in multiple ways, but only a portion of these descriptions are covered by the safety filter. Compared to token replacement strategies, our LLM-piloted multi-agent method can explore a larger semantically equivalent space more efficiently, owing to the LLM backbone’s advanced comprehension, generation, and instruction-following capabilities.

\noindent\textbf{Safety Implications:} Our method illustrated in Algorithm~\ref{alg-workflow} does not require online querying of the target T2I model during adversarial prompt generation. Moreover, as shown in our cost evaluation in \S\ref{sec-eval-cost}, generating an effective adversarial prompt is inexpensive, and these prompts can be reused multiple times for image generation as indicated in \S\ref{sec-reuse-bypass}. With the ongoing evolution of agents' backbone LLMs, the same cost will likely enable access to even more powerful models, making this an increasingly significant threat.

\noindent\textbf{Countermeasures:}
A possible defense is to apply post-generation safety filter on generated images, using vision understanding models or multi-modal foundation models to detect whether the image itself contains sensitive content. However, compared to text-level scrutiny, image understanding generally incurs higher costs and delays, which could hinder its widespread adoption in practice.
Another potential defense is prompt summarization. Our method generally expands the drawing prompt to have a more verbose version. Conversely, we could summarize these verbose adversarial prompts for screening. Based on our empirical tests, the summarized adversarial prompts still bypass safety filters with over 95\% success rate, although certain nuanced visual details may be lost due to summarization. Moreover, the sentence-by-sentence prompt feeding method shown in Figure~\ref{fig-generated_image_samples} (3) render summarization-based defenses less effective, as the adversarial content is introduced gradually, making it more challenging to detect. We plan to systematically study the impact of various summarization techniques in our future work.

\noindent\textbf{Evaluation with More Fine-grained Image Ontology:}
In our evaluation, we observed that generated images related to violence, crime, and discrimination align better with the original sensitive prompt compared to other two sensitive categories. This can be attributed to the granularity of the image ontology in our current implementation, as shown in Figure~\ref{fig-ontology}. Images depicting bloodiness and eroticism often include more detailed sensitive elements, such as blood, which were not thoroughly decomposed in our specified ontology. In contrast, for violence, dividing the description between the performer and recipient of the action effectively conceals sensitive semantics. 
In the future, we will explore a more fine-grained ontology specification to potentially improve attack effectiveness across a broader range of categories and against more T2I models. 

\noindent\textbf{Ethical Considerations:} We have responsibly disclosed our findings to relevant stakeholders. We hope our work will inspire positive applications, such as using our method as a red teaming tool to efficiently identify vulnerabilities.

\section{Conclusion}

In this study, we approach adversarial prompts against T2I models from a different perspective. To bypass both semantic safe/unsafe and logical/illogical checks, we design LLM-piloted method that rephrases sensitive prompts into adversarial versions considered safe and logically coherent, enabling the generation of images with intended visual effects.
Specifically, we design three agents, $\mathsf{Decomposer}$, $\mathsf{Polisher}$ and $\mathsf{Assembler}$, and use a specified image ontology to guide their workflow. Together, the agents isolates key visual components from targeted drawing prompts, articulate these components in benign terms, and reassemble them into a safe drawing prompt that objectively describes the appearance of visual components, effectively bypassing safety filters.
To evaluate our attack's capacity, we curated a prompt dataset covering 5 major topics censored by the latest T2I models, comprising 100 sensitive prompts and 3,600 corresponding adversarial prompts.
Our evaluation demonstrates that our method is both attack-effective and cost-effective. 
Our adversarial prompts can successfully bypass close-box safety filters of DALL·E 3 and Midjourney. 
With just 1 dollar, we can generate 28 adversarial prompts using GPT-4 as the agent backbone. 
Our findings highlight non-negligible safety implications, and we hope our open-sourced code and dataset facilitate future research.

{
    \small
    \bibliographystyle{ieeenat_fullname}
    \newpage
    \bibliography{main}

\begin{thebibliography}{51}
\providecommand{\natexlab}[1]{#1}
\providecommand{\url}[1]{\texttt{#1}}
\expandafter\ifx\csname urlstyle\endcsname\relax
  \providecommand{\doi}[1]{doi: #1}\else
  \providecommand{\doi}{doi: \begingroup \urlstyle{rm}\Url}\fi

\bibitem[are()]{arena}
{Chatbot Arena}.
\newblock \url{https://huggingface.co/spaces/lmsys/chatbot-arena-leaderboard}.

\bibitem[cha({\natexlab{a}})]{chatglm_pricing}
{ChatGLM API Pricing}.
\newblock \url{https://open.bigmodel.cn/dev/api\#product-billing}, {\natexlab{a}}.

\bibitem[cha({\natexlab{b}})]{chatgpt}
{OpenAI} {ChatGPT}.
\newblock \url{https://chat.openai.com/}, {\natexlab{b}}.

\bibitem[cli()]{clip_repo}
{CLIP Repo}.
\newblock \url{https://github.com/openai/CLIP}.

\bibitem[dal({\natexlab{a}})]{dalle2}
{DALL-E} 2.
\newblock \url{https://openai.com/dall-e-2}, {\natexlab{a}}.

\bibitem[dal({\natexlab{b}})]{dalle2_moderation}
{DALL-E} 2 moderation policy.
\newblock \url{https://openai.com/policies/usage-policies/}, {\natexlab{b}}.

\bibitem[dal({\natexlab{c}})]{dalle3}
{DALL-E} 3.
\newblock \url{https://openai.com/dall-e-3}, {\natexlab{c}}.

\bibitem[dal({\natexlab{d}})]{dalle_systemcard}
{DALL·E 3 system card}.
\newblock \url{https://openai.com/research/dall-e-3-system-card}, {\natexlab{d}}.

\bibitem[eth()]{ethical_concern}
{The Rise of Ethical Concerns about AI Content Creation: A Call to Action}.
\newblock \url{https://www.computer.org/publications/tech-news/trends/ethical-concerns-on-ai-content-creation}.

\bibitem[gpt()]{gpt4}
{GPT-4}.
\newblock \url{https://openai.com/research/gpt-4}.

\bibitem[mid({\natexlab{a}})]{midjourney}
Midjourney.
\newblock \url{https://www.midjourney.com/}, {\natexlab{a}}.

\bibitem[mid({\natexlab{b}})]{midjourney-ban}
{Midjourney's banned words policy}.
\newblock \url{https://openaimaster.com/midjourney-banned-words/}, {\natexlab{b}}.

\bibitem[ope({\natexlab{a}})]{openai_pricing}
{OpenAI API Pricing}.
\newblock \url{https://openai.com/pricing}, {\natexlab{a}}.

\bibitem[ope({\natexlab{b}})]{opencompass}
{OpenCompass}.
\newblock \url{https://opencompass.org.cn/leaderboard-llm}, {\natexlab{b}}.

\bibitem[qia({\natexlab{a}})]{qianwen_14_pricing}
{TongYiQianWen 14B API Pricing}.
\newblock \url{https://help.aliyun.com/zh/dashscope/developer-reference/tongyi-qianwen-7b-14b-72b-metering-and-billing?spm=a2c4g.11186623.0.0.693c502fDRJtKO}, {\natexlab{a}}.

\bibitem[qia({\natexlab{b}})]{qianwen_max_pricing}
{TongYiQianWen Max API Pricing}.
\newblock \url{https://help.aliyun.com/zh/dashscope/developer-reference/tongyi-thousand-questions-metering-and-billing?spm=a2c4g.11186623.0.i1}, {\natexlab{b}}.

\bibitem[spa()]{spark_pricing}
{Spark API Pricing}.
\newblock \url{https://www.xfyun.cn/doc/spark/Web.html}.

\bibitem[sup()]{superclueai}
{SuperCLUE}.
\newblock \url{https://www.superclueai.com/}.

\bibitem[nsf(2023)]{nsfwgpt2023}
Nsfw gpt.
\newblock \url{https://www.reddit.com/r/ChatGPT/comments/11vlp7j/nsfwgpt_that_nsfw_prompt/}, 2023.

\bibitem[Ba et~al.(2023)Ba, Zhong, Lei, Cheng, Wang, Qin, Wang, and Ren]{ba2023surrogateprompt}
Zhongjie Ba, Jieming Zhong, Jiachen Lei, Peng Cheng, Qinglong Wang, Zhan Qin, Zhibo Wang, and Kui Ren.
\newblock Surrogateprompt: Bypassing the safety filter of text-to-image models via substitution.
\newblock \emph{arXiv preprint arXiv:2309.14122}, 2023.

\bibitem[Bansal et~al.(2022)Bansal, Yin, Monajatipoor, and Chang]{bansal2022well}
Hritik Bansal, Da Yin, Masoud Monajatipoor, and Kai-Wei Chang.
\newblock How well can text-to-image generative models understand ethical natural language interventions?
\newblock In \emph{Proceedings of the 2022 Conference on Empirical Methods in Natural Language Processing}, pages 1358--1370, 2022.

\bibitem[Brown et~al.(2020)Brown, Mann, Ryder, Subbiah, Kaplan, Dhariwal, Neelakantan, Shyam, Sastry, Askell, et~al.]{brown2020language}
Tom Brown, Benjamin Mann, Nick Ryder, Melanie Subbiah, Jared~D Kaplan, Prafulla Dhariwal, Arvind Neelakantan, Pranav Shyam, Girish Sastry, Amanda Askell, et~al.
\newblock Language models are few-shot learners.
\newblock \emph{Advances in neural information processing systems}, 33:\penalty0 1877--1901, 2020.

\bibitem[Carlini and Wagner(2017)]{carlini2017towards}
Nicholas Carlini and David Wagner.
\newblock Towards evaluating the robustness of neural networks.
\newblock In \emph{2017 IEEE Symposium on Security and Privacy (SP)}, pages 39--57. IEEE, 2017.

\bibitem[Chen et~al.(2020)Chen, Huang, Huang, Zhang, and Wang]{chen2020ecgadv}
Huangxun Chen, Chenyu Huang, Qianyi Huang, Qian Zhang, and Wei Wang.
\newblock Ecgadv: Generating adversarial electrocardiogram to misguide arrhythmia classification system.
\newblock In \emph{Proceedings of the AAAI Conference on Artificial Intelligence}, pages 3446--3453, 2020.

\bibitem[Ganguli et~al.(2023)Ganguli, Askell, Schiefer, Liao, Luko{\v{s}}i{\=u}t{\.e}, Chen, Goldie, Mirhoseini, Olsson, Hernandez, et~al.]{ganguli2023capacity}
Deep Ganguli, Amanda Askell, Nicholas Schiefer, Thomas Liao, Kamil{\.e} Luko{\v{s}}i{\=u}t{\.e}, Anna Chen, Anna Goldie, Azalia Mirhoseini, Catherine Olsson, Danny Hernandez, et~al.
\newblock The capacity for moral self-correction in large language models.
\newblock \emph{arXiv e-prints}, pages arXiv--2302, 2023.

\bibitem[Garg and Ramakrishnan(2020)]{garg2020bae}
Siddhant Garg and Goutham Ramakrishnan.
\newblock Bae: Bert-based adversarial examples for text classification.
\newblock In \emph{Proceedings of the 2020 Conference on Empirical Methods in Natural Language Processing (EMNLP)}, pages 6174--6181, 2020.

\bibitem[George(2020)]{george2020nsfw}
R. George.
\newblock Nsfw words list on github, 2020.

\bibitem[Goodfellow et~al.(2014)Goodfellow, Shlens, and Szegedy]{goodfellow2014explaining}
Ian~J Goodfellow, Jonathon Shlens, and Christian Szegedy.
\newblock Explaining and harnessing adversarial examples.
\newblock \emph{arXiv preprint arXiv:1412.6572}, 2014.

\bibitem[Gozalo-Brizuela and Garrido-Merch{\'a}n(2023)]{gozalo2023survey}
Roberto Gozalo-Brizuela and Eduardo~C Garrido-Merch{\'a}n.
\newblock A survey of generative ai applications.
\newblock \emph{arXiv preprint arXiv:2306.02781}, 2023.

\bibitem[Han et~al.(2020)Han, Hu, Foschini, Chinitz, Jankelson, and Ranganath]{han2020deep}
Xintian Han, Yuxuan Hu, Luca Foschini, Larry Chinitz, Lior Jankelson, and Rajesh Ranganath.
\newblock Deep learning models for electrocardiograms are susceptible to adversarial attack.
\newblock \emph{Nature medicine}, 26\penalty0 (3):\penalty0 360--363, 2020.

\bibitem[Ho et~al.(2020)Ho, Jain, and Abbeel]{ho2020denoising}
Jonathan Ho, Ajay Jain, and Pieter Abbeel.
\newblock Denoising diffusion probabilistic models.
\newblock \emph{Advances in neural information processing systems}, 33:\penalty0 6840--6851, 2020.

\bibitem[Jin et~al.(2020)Jin, Jin, Zhou, and Szolovits]{jin2020bert}
Di Jin, Zhijing Jin, Joey~Tianyi Zhou, and Peter Szolovits.
\newblock Is bert really robust? a strong baseline for natural language attack on text classification and entailment.
\newblock In \emph{Proceedings of the AAAI conference on artificial intelligence}, pages 8018--8025, 2020.

\bibitem[Kurakin et~al.(2018)Kurakin, Goodfellow, and Bengio]{kurakin2018adversarial}
Alexey Kurakin, Ian~J Goodfellow, and Samy Bengio.
\newblock Adversarial examples in the physical world.
\newblock In \emph{Artificial Intelligence Safety and Security}, pages 99--112. Chapman and Hall/CRC, 2018.

\bibitem[Li et~al.(2018)Li, Ji, Du, Li, and Wang]{li2018textbugger}
Jinfeng Li, Shouling Ji, Tianyu Du, Bo Li, and Ting Wang.
\newblock Textbugger: Generating adversarial text against real-world applications.
\newblock \emph{arXiv preprint arXiv:1812.05271}, 2018.

\bibitem[Li(2022)]{li2022nsfw}
M. Li.
\newblock Nsfw text classifier on hugging face, 2022.

\bibitem[Ma et~al.(2024)Ma, Pang, Guo, Wei, and Guo]{ma2024coljailbreak}
Yizhuo Ma, Shanmin Pang, Qi Guo, Tianyu Wei, and Qing Guo.
\newblock Coljailbreak: Collaborative generation and editing for jailbreaking text-to-image deep generation.
\newblock In \emph{Proceedings of the 38th International Conference and Workshop on Neural Information Processing Systems}, 2024.

\bibitem[Markov et~al.(2023)Markov, Zhang, Agarwal, Nekoul, Lee, Adler, Jiang, and Weng]{markov2023holistic}
Todor Markov, Chong Zhang, Sandhini Agarwal, Florentine~Eloundou Nekoul, Theodore Lee, Steven Adler, Angela Jiang, and Lilian Weng.
\newblock A holistic approach to undesired content detection in the real world.
\newblock In \emph{Proceedings of the AAAI Conference on Artificial Intelligence}, pages 15009--15018, 2023.

\bibitem[Maus et~al.(2023)Maus, Chao, Wong, and Gardner]{maus2023adversarial}
Natalie Maus, Patrick Chao, Eric Wong, and Jacob Gardner.
\newblock Adversarial prompting for black box foundation models.
\newblock \emph{arXiv preprint arXiv:2302.04237}, 2023.

\bibitem[Mehrotra et~al.(2024)Mehrotra, Zampetakis, Kassianik, Nelson, Anderson, Singer, and Karbasi]{mehrotra2024tree}
Anay Mehrotra, Manolis Zampetakis, Paul Kassianik, Blaine Nelson, Hyrum Anderson, Yaron Singer, and Amin Karbasi.
\newblock Tree of attacks: Jailbreaking black-box llms automatically.
\newblock In \emph{Proceedings of the 38th International Conference and Workshop on Neural Information Processing Systems}, 2024.

\bibitem[Milli{\`e}re(2022)]{milliere2022adversarial}
Rapha{\"e}l Milli{\`e}re.
\newblock Adversarial attacks on image generation with made-up words.
\newblock \emph{arXiv preprint arXiv:2208.04135}, 2022.

\bibitem[Rahutomo et~al.(2012)Rahutomo, Kitasuka, Aritsugi, et~al.]{rahutomo2012semantic}
Faisal Rahutomo, Teruaki Kitasuka, Masayoshi Aritsugi, et~al.
\newblock Semantic cosine similarity.
\newblock In \emph{The 7th international student conference on advanced science and technology ICAST}, page~1. University of Seoul South Korea, 2012.

\bibitem[Rombach et~al.(2022)Rombach, Blattmann, Lorenz, Esser, and Ommer]{rombach2022high}
Robin Rombach, Andreas Blattmann, Dominik Lorenz, Patrick Esser, and Bj{\"o}rn Ommer.
\newblock High-resolution image synthesis with latent diffusion models.
\newblock In \emph{Proceedings of the IEEE/CVF conference on computer vision and pattern recognition}, pages 10684--10695, 2022.

\bibitem[Saharia et~al.(2022)Saharia, Chan, Saxena, Li, Whang, Denton, Ghasemipour, Gontijo~Lopes, Karagol~Ayan, Salimans, et~al.]{saharia2022photorealistic}
Chitwan Saharia, William Chan, Saurabh Saxena, Lala Li, Jay Whang, Emily~L Denton, Kamyar Ghasemipour, Raphael Gontijo~Lopes, Burcu Karagol~Ayan, Tim Salimans, et~al.
\newblock Photorealistic text-to-image diffusion models with deep language understanding.
\newblock \emph{Advances in Neural Information Processing Systems}, 35:\penalty0 36479--36494, 2022.

\bibitem[Sanh et~al.(2019)Sanh, Debut, Chaumond, and Wolf]{sanh2019distilbert}
Victor Sanh, Lysandre Debut, Julien Chaumond, and Thomas Wolf.
\newblock Distilbert, a distilled version of bert: smaller, faster, cheaper and lighter.
\newblock \emph{arXiv preprint arXiv:1910.01108}, 2019.

\bibitem[Shan et~al.(2023)Shan, Ding, Passananti, Zheng, and Zhao]{shan2023promptspecific}
Shawn Shan, Wenxin Ding, Josephine Passananti, Haitao Zheng, and Ben~Y. Zhao.
\newblock Prompt-specific poisoning attacks on text-to-image generative models.
\newblock \emph{arXiv preprint arXiv:2310.13828}, 2023.

\bibitem[Vaswani et~al.(2017)Vaswani, Shazeer, Parmar, Uszkoreit, Jones, Gomez, Kaiser, and Polosukhin]{vaswani2017attention}
Ashish Vaswani, Noam Shazeer, Niki Parmar, Jakob Uszkoreit, Llion Jones, Aidan~N Gomez, {\L}ukasz Kaiser, and Illia Polosukhin.
\newblock Attention is all you need.
\newblock \emph{Advances in neural information processing systems}, 30, 2017.

\bibitem[Wei et~al.(2022)Wei, Wang, Schuurmans, Bosma, Xia, Chi, Le, Zhou, et~al.]{wei2022chain}
Jason Wei, Xuezhi Wang, Dale Schuurmans, Maarten Bosma, Fei Xia, Ed Chi, Quoc~V Le, Denny Zhou, et~al.
\newblock {Chain-of-Thought Prompting Elicits Reasoning in Large Language Models}.
\newblock \emph{Advances in neural information processing systems}, 35:\penalty0 24824--24837, 2022.

\bibitem[Xue et~al.(2024)Xue, Zheng, Hua, Shen, Liu, B{\"o}l{\"o}ni, and Lou]{xue2024trojllm}
Jiaqi Xue, Mengxin Zheng, Ting Hua, Yilin Shen, Yepeng Liu, Ladislau B{\"o}l{\"o}ni, and Qian Lou.
\newblock Trojllm: A black-box trojan prompt attack on large language models.
\newblock \emph{Advances in Neural Information Processing Systems}, 36, 2024.

\bibitem[Yang et~al.(2023)Yang, Hui, Yuan, Gong, and Cao]{yang2023sneakyprompt}
Yuchen Yang, Bo Hui, Haolin Yuan, Neil Gong, and Yinzhi Cao.
\newblock Sneakyprompt: Evaluating robustness of text-to-image generative models' safety filters.
\newblock \emph{arXiv preprint arXiv:2305.12082}, 2023.

\bibitem[Zhu et~al.(2023)Zhu, Wang, Zhou, Wang, Chen, Wang, Yang, Ye, Zhang, Zhenqiang~Gong, et~al.]{zhu2023promptbench}
Kaijie Zhu, Jindong Wang, Jiaheng Zhou, Zichen Wang, Hao Chen, Yidong Wang, Linyi Yang, Wei Ye, Yue Zhang, Neil Zhenqiang~Gong, et~al.
\newblock Promptbench: Towards evaluating the robustness of large language models on adversarial prompts.
\newblock \emph{arXiv e-prints}, pages arXiv--2306, 2023.

\bibitem[Zou et~al.(2023)Zou, Wang, Carlini, Nasr, Kolter, and Fredrikson]{zou2023universal}
Andy Zou, Zifan Wang, Nicholas Carlini, Milad Nasr, J~Zico Kolter, and Matt Fredrikson.
\newblock Universal and transferable adversarial attacks on aligned language models.
\newblock \emph{arXiv preprint arXiv:2307.15043}, 2023.

\end{thebibliography}
}
\newpage


\end{document}